\documentclass[pmlr]{jmlr}

\RequirePackage{graphicx}
 \usepackage{booktabs}
\usepackage{longtable}
\makeatletter
\def\set@curr@file#1{\def\@curr@file{#1}} 
\makeatother
\usepackage[load-configurations=version-1]{siunitx} 

\theorembodyfont{\upshape}
\theoremheaderfont{\scshape}
\theorempostheader{:}
\theoremsep{\newline}

\jmlryear{2024}
\jmlrworkshop{Machine Learning for Healthcare}

\title[Onco-Retriever]{Onco-Retriever: Generative Classifier for Retrieval of EHR Records in Oncology}

\author{\Name{Aditya Basu*}
       \Email{aditya.basu@triomics.in}\\ 
       \addr
       Triomics Research\\
       San Francisco, California, USA
       \AND
        \Name{Shashi Gupta*}
       \Email{shashi.gupta@triomics.in}\\ 
       \addr
       Triomics Research\\
       San Francisco, California, USA
       \AND
        \Name{Bradley Taylor}
       \Email{btaylor@mcw.edu}\\ 
       \addr
       Medical College of Wisconsin\\
       Milwaukee, USA 
       \AND
       \Name{Anai Kothari}
       \Email{akothari@mcw.edu}\\ 
       \addr
       Medical College of Wisconsin\\
       Milwaukee, Wisconsin, USA 
       \AND
       \Name{Hrituraj Singh}
       \Email{hrituraj@triomics.com}\\ 
       \addr
       Triomics Research\\
       San Francisco, California, USA 
}
\begin{document}

\maketitle

\begin{abstract}
Retrieving information from EHR systems is essential for answering specific questions about patient journeys and improving the delivery of clinical care. Despite this fact, most EHR systems still rely on keyword-based searches. With the advent of generative large language models (LLMs), retrieving information can lead to better search and summarization capabilities. Such retrievers can also feed Retrieval-augmented generation (RAG) pipelines to answer any query. However, the task of retrieving information from EHR real-world clinical data contained within EHR systems in order to solve several downstream use cases is challenging due to the difficulty in creating query-document support pairs. We provide a blueprint for creating such datasets in an affordable manner using large language models. Our method results in a retriever that is 30-50 F-1 points better than propriety counterparts such as Ada and Mistral for oncology data elements. We further compare our model, called Onco-Retriever, against fine-tuned PubMedBERT models as well. We conduct extensive manual evaluation along with latency analysis of the different models and provide a path forward for healthcare organizations to build domain-specific retrievers. All our experiments were conducted on real patient EHR data.
\end{abstract}

\section{Introduction}

The extensive data present in Electronic Health Records (EHRs) is predominantly unstructured, posing significant challenges in healthcare data management (\cite{unstructured1, unstructured2}). The sheer volume and complexity of this data hinder efficient summarization of patient journeys (\cite{summarization1, summarization2, summarization3}), impede the search for pertinent information (\cite{search1, search2, search3, search4}), and complicate the task of answering critical questions, such as those necessary for clinical trial matching (\cite{ct1, ct2, ct3}). This not only delays care delivery but also contributes to clinical burnout, as healthcare professionals end up spending considerable time navigating through EHRs to locate relevant data (\cite{burden1, burden2, burden3, burden4, burden5}).

Despite the strides made in natural language processing with instruction-tuned language models, a significant gap remains in their application to effective information retrieval (\cite{gpt3, gpt4, chatgpt, mistral, llama2, triplets}). Current methodologies largely rely on embedding similarity techniques (\cite{embedding1, embedding2}). Although popular, these methods often result in suboptimal performance when applied to clinical datasets due to the unique features of clinical and biomedical text (\cite{medretriever1, medretriever2}). The inherent limitation lies in the predominant use of smaller BERT-like models in embedding-based systems. Such models, while efficient, often fail to capture the depth and specificity required for cross-domain applicability. One of the primary challenges in this domain is the scarcity of query-passage annotations, especially pertinent to EHR data. This hampers the development of effective retrieval systems. 

Furthermore, while there have been concerted efforts in training specialized biomedical models (\cite{pubmedbert, medretriever1}), including biomedical retrievers, the specific task of creating retrievers tailored for clinical EHR data remains significantly underexplored. Even within the clinical data domain, though there has been progress in developing embedding models for tasks like named entity recognition and information extraction, the work on retrieval is limited (\cite{pubmedbert, scibert, domainmodel1, domainmodel2}). Improving EHR data retrieval is a unique opportunity; retrieval processes for EHR data is predominantly centered around standard and fixed concepts, particularly in specialized fields such as oncology. Capitalizing on this, our approach to building a retriever is tailored to focus on these high-priority concepts, therefore aligning the retrieval mechanism to the clinical need for EHR data summarization in oncological contexts.

In this study, we introduce a lightweight, cost-effective methodology for creating oncology-specific retrievers from the ground up. Our investigation is concentrated on 13 principal concepts of interest to oncologists based on commonalities across eight vocabularies that represent cancer health information (ICD-O-3, NCIt, NAACR, CAP, AJCC, HemOnc, ATC, mCODE). Our contributions can be summarized as follows -

\begin{itemize}
  \item \textbf{Synthetic Dataset Generation:} Utilizing GPT-3, we successfully generate a synthetic dataset from real world EHR data specifically designed to facilitate the distillation process of creating a specialized retriever.
  \item \textbf{Development of a Compact Onco-Retriever Model:} The distilled retriever model, with fewer than 500 million parameters, not only is considerably smaller but also surpasses the performance of conventional embedding-based models, including GPT-3.
  \item \textbf{Local Deployment Capability:} Owing to its reduced size, the model is adept for local deployment, addressing a critical need in healthcare for data security and privacy.
  \item \textbf{Extensive Model Evaluation:} We conduct comprehensive evaluations, focusing on incrementally increasing the size of the distilled model while concurrently enhancing its performance through manual annotations and targeted evaluation strategies.
  \item \textbf{Latency Benchmarking:} The latency of the model is thoroughly benchmarked, demonstrating its readiness for integration into production environments, an essential consideration for real-world healthcare applications.
\end{itemize}

\begin{figure}[ht]
  \centering
  \includegraphics[width=0.85\textwidth]{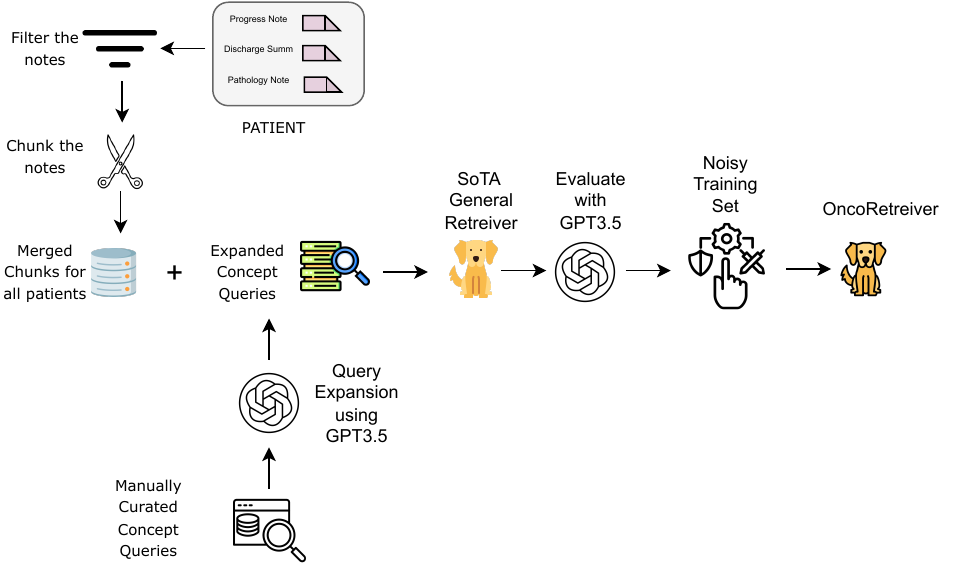}
  \caption{Patient notes are first filtered and chunked to remove unnecessary information in the dataset. The merged chunks for all the patients in the training set are then passed through a generic retriever on the relevant expanded query set that is manually curated. Retrieved top-k chunks are then evaluated one chunk at a time using GPT-3.5. The evaluation output is generated in a way that it results in a training dataset. The retriever model is then finetuned on this set to train the onco-retriever.}
  \label{fig:retreiver}
\end{figure}

\section{Related Work}
Retrieving information from Electronic Health Records (EHR) predominantly depends on keyword-based search, a method that, while prevalent, is increasingly recognized for its limitations (\cite{search1, search2, search3}). To enhance these systems, there has been a move towards integrating taxonomy graphs, which aim to improve the accuracy and relevance of retrieved information (\cite{emerse}). Such efforts can be much more helpful than standard keyword-based approaches (\cite{emerse2}). 

However, the development and implementation of such taxonomy-augmented systems demands extensive effort and planning. The challenge is further compounded by the need to encompass a wide array of terminologies and concepts, which are often variably documented in  EHR notes that are created for the purpose of clinical care. This variability poses a significant hurdle; even with well-structured taxonomies, it is daunting, if not impossible, to capture every possible permutation in which a concept might be mentioned. Such techniques can be replaced using more sophisticated semantic retrieval engines.

The majority of existing semantic retrieval solutions predominantly leverage the SBERT framework, utilizing BERT-based embeddings to align sentences with similar meanings within the embedding space (\cite{embedding1, embedding2}). The interpretation of 'similarity' in these methods is contingent upon the dataset used, leading to variability in its definition (\cite{triplets, semanticembedding1, semanticembedding2}). Such popular approaches have resulted in the emergence of numerous open-source (\cite{mpnet,minilm}) and proprietary embedding-based retrieval models(\cite{ada}) . Recent research has also demonstrated the feasibility of pruning generative models for information retrieval purposes(\cite{mistralsfr}). However, these models often conceptualize 'similarity' differently, a notion not entirely aligned with the requirements of healthcare data. Furthermore, for scalability across large text collections, these models are typically more compact, which in turn affects their effectiveness in domain transfer. This becomes clear when comparing the decline in performance of embedding-based models to that of larger models like GPT-4 when transitioning from general to healthcare data. While GPT4 still performs reasonably when moving from public data to healthcare data due it large size and robustness, smaller models are more likely to fail during domain transfer.

Recent developments in the field of biomedical and clinical natural language processing have seen a significant push towards the creation of encoder models specifically designed for healthcare data, with notable examples including PubMedBERT, SciBERT, ClinicalBERT, and CancerBERT (\cite{pubmedbert, scibert, clinicalbert, cancerbert}). These models are predominantly focused on tasks like entity extraction, which limits their direct applicability in retrieval-based frameworks due to their inherent need for fine-tuning on labeled data. 

To address this gap, there have been initiatives to develop fine-tuned retriever models built upon these encoder frameworks (\cite{medretriever1, medretriever2}). However, these models have primarily been tailored for biomedical, rather than EHR data.

One of the primary challenges in adapting these systems is the need for parallel query-passage datasets for the target dataset (\cite{msmarco, nli, dpr}). However, it is extremely difficult to curate or create such data in healthcare unless a proxy dataset is modified to serve the needs. Our work shows that this can be readily achieved by using the power of generalizable large language models.

\section{Methods}

\subsection{Problem Formulation}
Our objective is to formulate and solve the task of extracting pertinent information chunks from patient journeys documented in the EHR, focusing on predefined key concepts. We define this retrieval problem mathematically as follows:

Let \( N = \{n_1, n_2, ..., n_m\} \) be the set of patient notes, where each note \( n_i \) is an individual document. We employ a rule-based chunking algorithm to transform each \( n_i \) into a set of chunks \( C_i = \{c_1, c_2, ..., c_t\} \). Thus, for the entire patient journey, we have a chunk collection \( C = \bigcup_{i=1}^{m} C_i \).

Given a set of predefined concepts \( \mathcal{K} \) as defined in Table \ref{tab:concept_definitions}, identified through consultations with domain experts, the retrieval task for a concept \( k \in \mathcal{K} \) is defined by two key metrics:

1. \textbf{Recall}: This is measured as the ratio of the number of relevant chunks for concept \( k \) retrieved in the top-k results to the total number of relevant chunks for \( k \) in the $C$. Mathematically, for a retrieval function \( R(N, k) \) that retrieves a set of chunks for \( k \) from \( N \), recall is defined as:
   \[ \text{Recall}(k) = \frac{|\{c \in R(N, k) \cap \text{Relevant}(k)\}|}{|\text{Relevant}(k)|} \]

2. \textbf{Precision}: This evaluates the proportion of the top-k retrieved chunks that are indeed true positives for concept \( k \). It is defined as:
   \[ \text{Precision}(k) = \frac{|\{c \in R(N, k) \cap \text{Relevant}(k)\}|}{|R(N, k)|} \]

Here, \( \text{Relevant}(k) \) denotes the set of chunks that are genuinely relevant to concept \( k \). Our goal is to maximize both Recall and Precision for each concept \( k \in \mathcal{K} \) across all patient notes in \( N \).

\begin{table}[h]
\centering
\begin{tabular}{|l|p{10cm}|}
\hline
\textbf{Concept Name}              & \textbf{Definition} \\ \hline
Current Diagnosis                  & The current confirmed diagnosis of the patient, indicating the origin and site of the cancer. \\ \hline
Disease Status                     & Status of the disease, including conditions like Locally Advanced, Resectable, or Metastasized, along with details on metastasis sites. \\ \hline
Tumor Characteristics              & Detailed histologic classification of cancer based on cellular structure and tissue origin, identified through microscopic examination, including tumor size and grade. \\ \hline
Tumor Staging                      & A combination of the size and extent of the primary tumor (T), lymph node involvement (N), and metastasis (M). \\ \hline
Combined Stage Grouping            & Combined evaluation based on tumor size, lymph node involvement, and metastasis, providing a comprehensive stage grouping. \\ \hline
Treatment Outcomes                 & Documented outcomes from treatments received, which may include remission, stable disease, or progression, etc. \\ \hline
Treatment Types                    & List of treatments administered, including types such as chemotherapy, hormone therapy, or immunotherapy. \\ \hline
Biomarkers Assessed                & Details of specific biomarkers tested to inform treatment decisions and prognosis, including genetic mutations. \\ \hline
Surgical Interventions             & Record of all surgical procedures the patient has undergone in relation to their cancer diagnosis. \\ \hline
Diagnostic Assessments             & Comprehensive records of diagnostic methods used to determine the specifics of the patient's cancer. \\ \hline
Diagnosis Date                     & The date when the patient's cancer was first confirmed through pathological diagnosis. \\ \hline
Family History                     & Cancer-related family history, noting instances of cancer in immediate and extended family members. \\ \hline
Scores                             & Performance-related scores such as ECOG, Karnofsky, along with other related scores. \\ \hline
\end{tabular}
\caption{Definitions of Oncology Retrieval Concepts}
\label{tab:concept_definitions}
\end{table}

\subsection{Baselines}
In order to create a strong baseline for our study, we utilized state-of-the-art embedding-based retrievers as outlined in \cite{embedding1, embedding2}. The process includes the following  steps:

\begin{enumerate}
  \item \textbf{Chunking}: As discussed in the previous section, given a set of notes \( N = \{n_1, n_2, ..., n_m\} \) from patient records, we partitioned each note \(n_i\) into a set of smaller chunks \(C_i = \{c_1, c_2, ..., c_m\}\). This process was important to enhance the manageability and retrieval efficiency of the data. It is also consistent with the standard practice.
  
  \item \textbf{Embedding and Distance Measurement}: Each query \(q_i\) and chunk \(c_j\) were embedded into a shared vector space, resulting in vectors \(v(q_i)\) and \(v(c_j)\). The cosine distance \(d(q_i, c_j)\) between these vectors was computed to quantify the similarity, where
      \[d(q_i, c_j) = 1 - \frac{v(q_i) \cdot v(c_j)}{\|v(q_i)\| \|v(c_j)\|}\]
  
  \item \textbf{Chunk Assignment}: The chunks were then ordered by their proximity to each query. Let \(D(q_i) = \{d(q_i, c_1), d(q_i, c_2), ..., d(q_i, c_m)\}\) be the set of distances from query \(q_i\) to all chunks. The chunks were sorted in ascending order of distances in \(D(q_i)\), associating the chunks with the concept. A chunk can include multiple concepts.
\end{enumerate}

To augment these baseline models, we further integrated the technique of query expansion (\cite{queryexpansion1, queryexpansion2, queryexpansion3, queryexpansion4}). For each concept \(k\) in our predefined list (see Table 1), we generated a set of 30 candidate queries \(Q_k = \{q_1, q_2, ..., q_{30}\}\) using GPT-4. These queries were specifically created to align with the typical representations of concept information in EHR records by designing a suitable prompt. We have added the formulated queries along with the prompt in the supplementary information.

\subsection{Onco-Retrievers}
\begin{figure}[ht]
  \centering
  \includegraphics[width=1\textwidth]{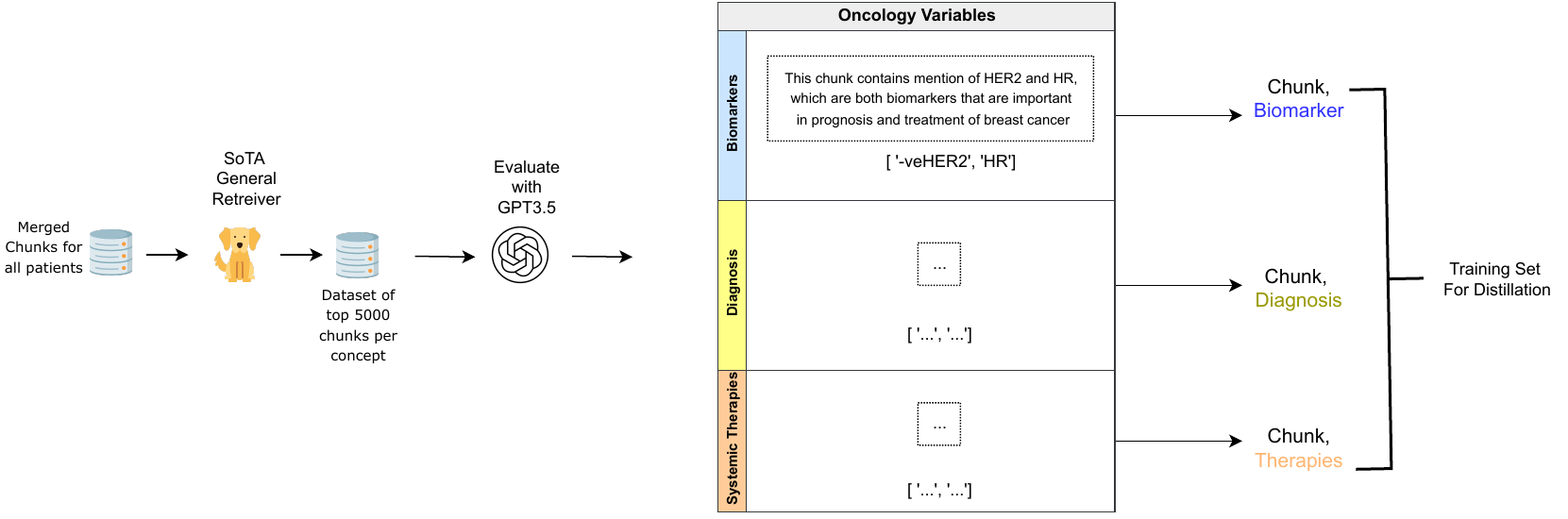}
  \caption{Out of our training set, we first use the state of the art retriever to find top 500 chunks. We use GPT3.5 to generate and label data for each of the chunks. This results in creation of multiple (chunk, concept) label pairs which we treat as training instances }
  \label{fig:training}
\end{figure}

We developed three variations of Onco-Retrievers: \textit{Small}, \textit{Optimized}, and \textit{Large}. The \textit{Small} and \textit{Optimized} models are derivatives of the Qwen1.5 500M parameter model \cite{qwen}, whereas the \textit{Large} variant is an extension of the Gemma model \cite{gemma}. The training procedure for these models involved several steps:

\begin{enumerate}
  \item Initial Chunk Filtering: Utilizing OpenAI's Ada model, we filtered out 5,000 chunks from our training set for each concept, resulting in a dataset of 65,000 chunks, agnostic of patient origin.
  
  \item Labeling via GPT-3.5: Each chunk was labeled using GPT-3.5, as outlined in the prompt provided in the appendix. The output, similar to Figure \ref{fig:training}, leveraged a Chain of Thought (CoT) approach to derive labels, enhancing model performance through better reasoning processes. This has proven very effective in improving the model performance [CITE]. 
  
  \item Model Training: With this dataset, the following models were trained:
    \begin{itemize}
      \item \textbf{Onco-Retriever (Small)}: A 500M parameter model, distilled on individual concepts. Formally, for each chunk \(c_i \in C\) and a total of \(\mathcal{K}\) concepts, the model trained on each concept \(k \in \mathcal{K}\) separately, resulting in a focused yet computationally intensive process.
      
      \item \textbf{Onco-Retriever (Optimized)}: Another 500M parameter model variant was fine-tuned to generate labels for all \(\mathcal{K}\) concepts simultaneously. This model was more efficient, albeit with a slight trade-off in accuracy.
      
      \item \textbf{Onco-Retriever (Large)}: Built on the Gemma model with 2B parameters, this variant mirrored the \textit{Optimized} model's approach but with significantly higher computational requirements.
    \end{itemize}
\end{enumerate}

The comparative effectiveness and computational trade-offs of these models are discussed in detail in the Results section.

Our training process is based on an informed assumption that output of GPT3.5, while, not fully accurate can still be used to distill smaller models that can perform even better than their original teacher model (\cite{distillinglmsbe, distillinglmspm}).

\textbf{PubMedBERT Comparison}: Alongside the Ada and Mistral baselines, our Onco-Retrievers were benchmarked against the fine-tuned version of PubMedBERT. PubMedBERT, a pre-trained language model specifically crafted for biomedical texts, has shown remarkable success in biomedical benchmarks. In our study, PubMedBERT served as a classifier, providing a basis to contrast the efficacy of Chain of Thought (CoT) based distillation against traditional classification techniques.

For the comparison, PubMedBERT was configured as a multi-label classifier. Given a set of labels \(L = \{l_1, l_2, ..., l_n\}\), corresponding to the \(\mathcal{K}\) concepts in our predefined list, the model was trained to predict multiple labels for each input chunk. The training objective was to minimize the classification loss across all labels for each chunk, formally represented as minimizing \( \sum_{i=1}^{n} \text{Loss}(c_i, L) \), where \(c_i\) is a chunk and \(L\) represents the set of its true labels.




\section{Dataset and Results} 

\subsection{Dataset}
We utilized real-world Electronic Health Records (EHRs) from 290 oncology patients of Medical College of Wisconsin EHR system, with each patient having an average of 200 documents. The patient records were split into two sets: a training set consisting of 240 patients and a separate test set of 50 patients.

To prepare the data for training, we employed a semantic chunking approach to split each document into contextually relevant and isolated chunks, ensuring that semantically related information remained together. From the training set chunks, we calculated the 5,000 chunks most semantically similar to each of the 13 predefined oncology concepts using cosine similarity with their text-ada-002 embeddings. This process yielded a total of 65,000 datapoints likely to contain relevant context for the targeted concepts.

To further refine the dataset, we used GPT-3 to extract relevant terms from each of the 65,000 datapoints, processing one concept at a time. Regex filtering was applied to minimize false positives, and GPT-3's self-verification capability was utilized to remove any false negatives. This resulted in a silver-annotation of each chunk for each concept.

For the golden test dataset, we ranked the document chunks from the 50 patients in the test set based on their similarity to each relevant concept. We selected 2 chunks with high similarity and 2 random chunks for each concept from every patient, resulting in a total of 2,600 chunks. 

These chunks were then annotated for the 13 concepts manually, providing us with a high-quality golden test set for evaluating the performance of the Onco-Retriever models.
\subsection{Results} 
The Onco-Retriever models (Small, Optimized, and Large) consistently outperformed the baseline models (Ada, Mistral, and PubMedBERT) in terms of precision and recall across the 13 oncology concepts studied.

\begin{figure}[ht]
  \centering
  \includegraphics[width=1\textwidth]{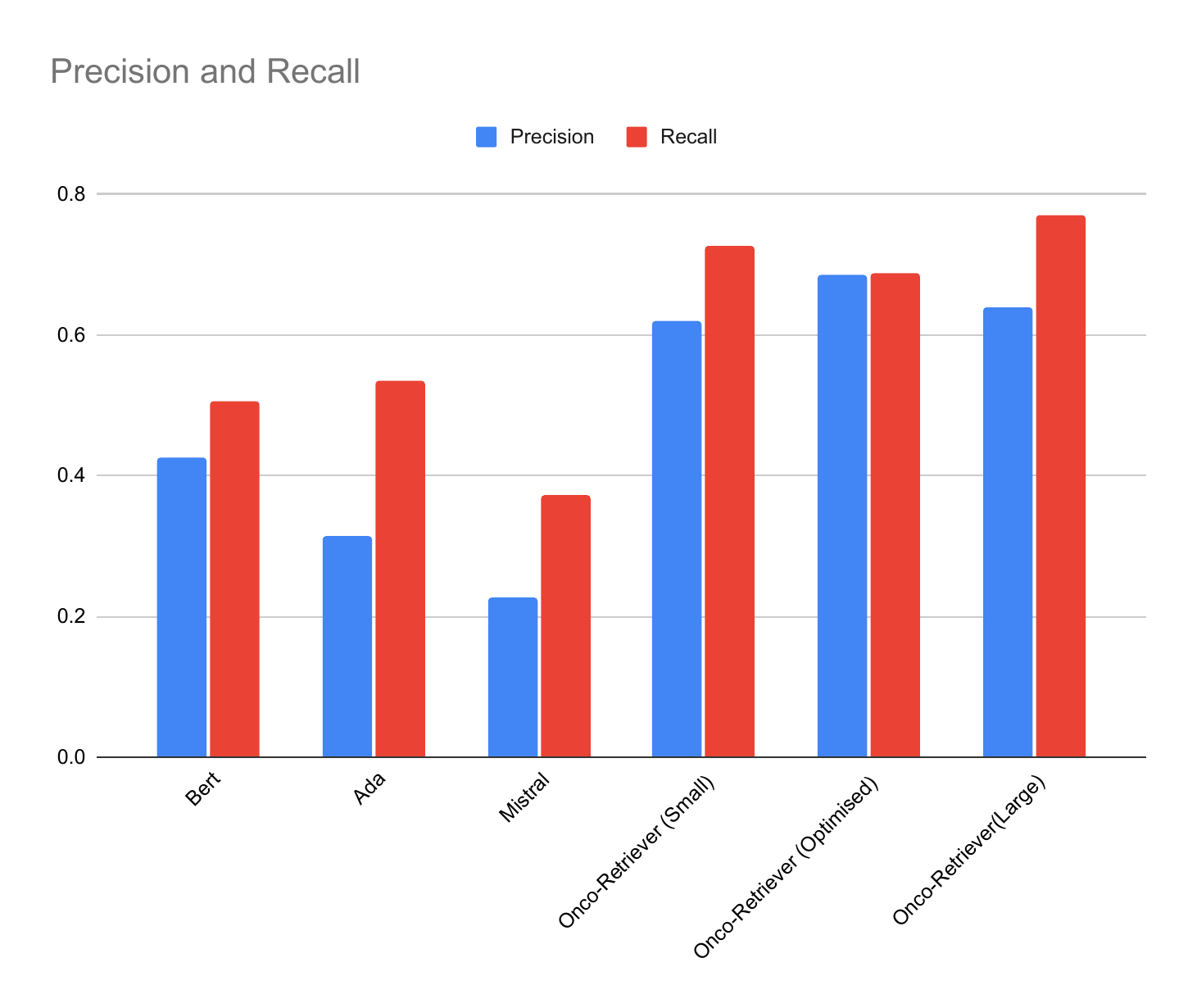}
  \caption{Precision and Recall Results across multiple retrievers }
  \label{fig:training}
\end{figure}


\begin{table}[]
\centering
\resizebox{\columnwidth}{!}{%
\begin{tabular}{lrrrrrrrrrrrr}
\textbf{Concept}       & \multicolumn{2}{l}{Onco-Ret (S)}  & \multicolumn{2}{l}{Onco-Ret (O)} & \multicolumn{2}{l}{PubmedBert}              & \multicolumn{2}{l}{Onco-Ret (L)}  & \multicolumn{2}{l}{Open AI Ada}                     & \multicolumn{2}{l}{Mistral SFR}                 \\
                       & p                    & r                    & p                      & r                     & p                    & r                    & p                    & r                    & p                    & r                    & p                    & r                    \\
Current Diagnosis      & 0.85                 & 0.86                 & 0.81                   & 0.78                  & 0.41                 & 0.50                 & 0.62                 & 0.75                 & 0.59                 & 0.46                 & 0.51                 & 0.39                 \\
Disease Status         & 0.65                 & 0.76                 & 0.68                   & 0.71                  & 0.41                 & 0.50                 & 0.61                 & 0.75                 & 0.40                 & 0.49                 & 0.26                 & 0.32                 \\
Tumor Characteristics  & 0.37                 & 0.77                 & 0.42                   & 0.72                  & 0.42                 & 0.56                 & 0.63                 & 0.83                 & 0.40                 & 0.56                 & 0.30                 & 0.41                 \\
Tumor Staging          & 0.74                 & 0.84                 & 0.69                   & 0.81                  & 0.47                 & 0.54                 & 0.71                 & 0.96                 & 0.29                 & 0.71                 & 0.24                 & 0.59                 \\
Combined Stage         & 0.52                 & 0.67                 & 0.69                   & 0.62                  & 0.39                 & 0.43                 & 0.58                 & 0.64                 & 0.28                 & 0.80                 & 0.13                 & 0.47                 \\
Treatment Outcomes     & 0.61                 & 0.69                 & 0.75                   & 0.69                  & 0.46                 & 0.48                 & 0.68                 & 0.72                 & 0.22                 & 0.52                 & 0.22                 & 0.52                 \\
Treatment Types        & 0.70                 & 0.73                 & 0.84                   & 0.77                  & 0.51                 & 0.57                 & 0.77                 & 0.85                 & 0.38                 & 0.39                 & 0.25                 & 0.26                 \\
Biomarkers Assessed    & 0.42                 & 0.63                 & 0.57                   & 0.58                  & 0.35                 & 0.45                 & 0.53                 & 0.67                 & 0.24                 & 0.55                 & 0.12                 & 0.27                 \\
Surgical Interventions & 0.64                 & 0.69                 & 0.78                   & 0.66                  & 0.36                 & 0.64                 & 0.54                 & 0.95                 & 0.35                 & 0.48                 & 0.34                 & 0.46                 \\
Diagnostic Assessments & 0.70                 & 0.76                 & 0.64                   & 0.74                  & 0.49                 & 0.49                 & 0.73                 & 0.73                 & 0.29                 & 0.31                 & 0.15                 & 0.17                 \\
Diagnosis Date         & 0.59                 & 0.72                 & 0.59                   & 0.76                  & 0.38                 & 0.36                 & 0.57                 & 0.54                 & 0.42                 & 0.52                 & 0.31                 & 0.38                 \\
Family History         & 0.62                 & 0.66                 & 0.70                   & 0.53                  & 0.44                 & 0.51                 & 0.66                 & 0.77                 & 0.08                 & 0.65                 & 0.05                 & 0.38                 \\
Scores                 & 0.64                 & 0.66                 & 0.77                   & 0.57                  & 0.45                 & 0.56                 & 0.67                 & 0.84                 & 0.13                 & 0.52                 & 0.05                 & 0.21                 \\
                       & \multicolumn{1}{l}{} & \multicolumn{1}{l}{} & \multicolumn{1}{l}{}   & \multicolumn{1}{l}{}  & \multicolumn{1}{l}{} & \multicolumn{1}{l}{} & \multicolumn{1}{l}{} & \multicolumn{1}{l}{} & \multicolumn{1}{l}{} & \multicolumn{1}{l}{} & \multicolumn{1}{l}{} & \multicolumn{1}{l}{} \\
\textbf{Overall}       & \textbf{0.62}        & \textbf{0.73}        & \textbf{0.69}          & \textbf{0.69}         & \textbf{0.43}        & \textbf{0.51}        & \textbf{0.64}        & \textbf{0.77}        & \textbf{0.31}        & \textbf{0.54}        & \textbf{0.23}        & \textbf{0.37}       
\end{tabular}%
}
\end{table}

Onco-Retriever (Small), with fewer than 500 million parameters, achieved an overall precision of 0.62 and recall of 0.73, surpassing the performance of Ada (precision: 0.31, recall: 0.54) and Mistral (precision: 0.23, recall: 0.37) by a significant margin. The Small variant demonstrated the highest precision for concepts such as Current Diagnosis (0.85), Tumor Staging (0.74), and Diagnostic Assessments (0.70). 
\begin{figure}[ht]
  \centering
  \includegraphics[width=1\textwidth]{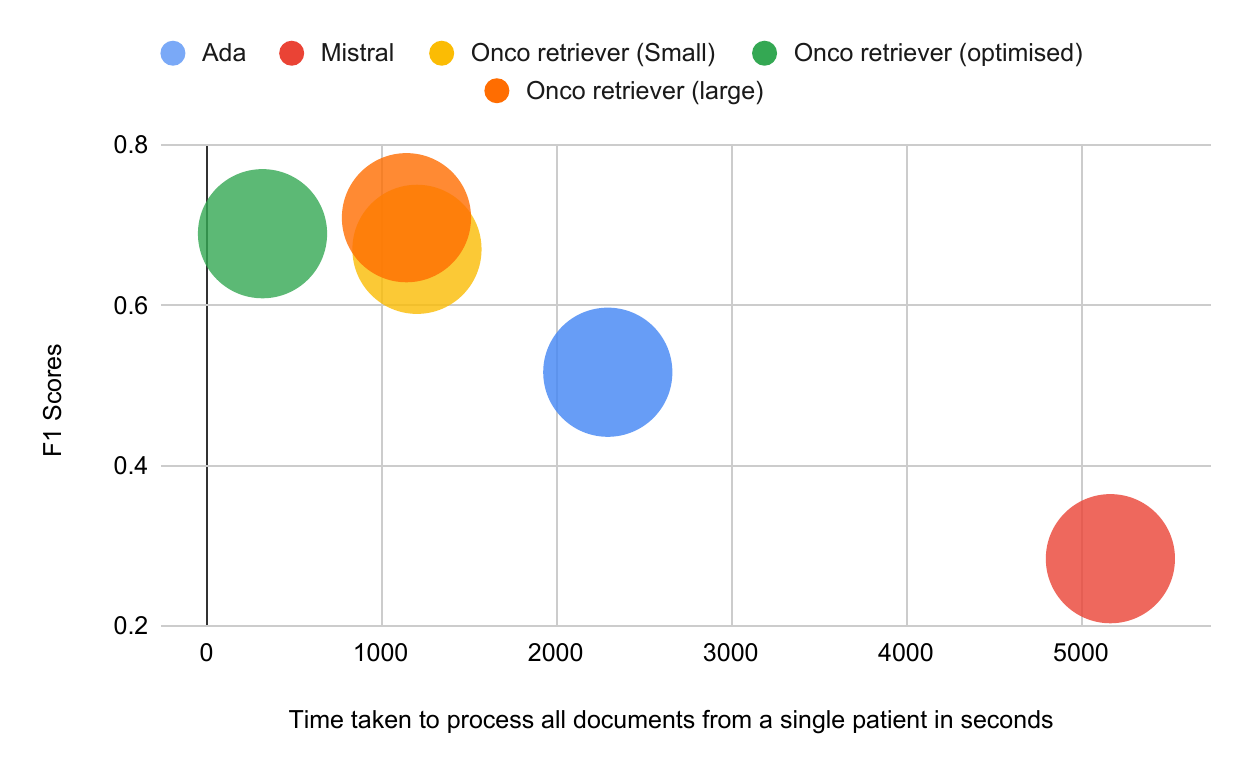}
  \caption{Average time taken by each model to process all the documents pertaining to a single patient vs F1 Scores fo}
  \label{fig:training}
\end{figure}

Onco-Retriever (Optimized), also with 500 million parameters, further improved the overall performance, with a precision of 0.69 and recall of 0.69. This variant showed the best precision for concepts like Treatment Types (0.84), Treatment Outcomes (0.75), and Family History (0.70). The Optimized model achieved these results while maintaining a significantly lower latency (318 seconds per patient) compared to the Small variant (1200 seconds per patient)
5.

\begin{figure}[ht]
  \centering

  \includegraphics[width=1\textwidth]{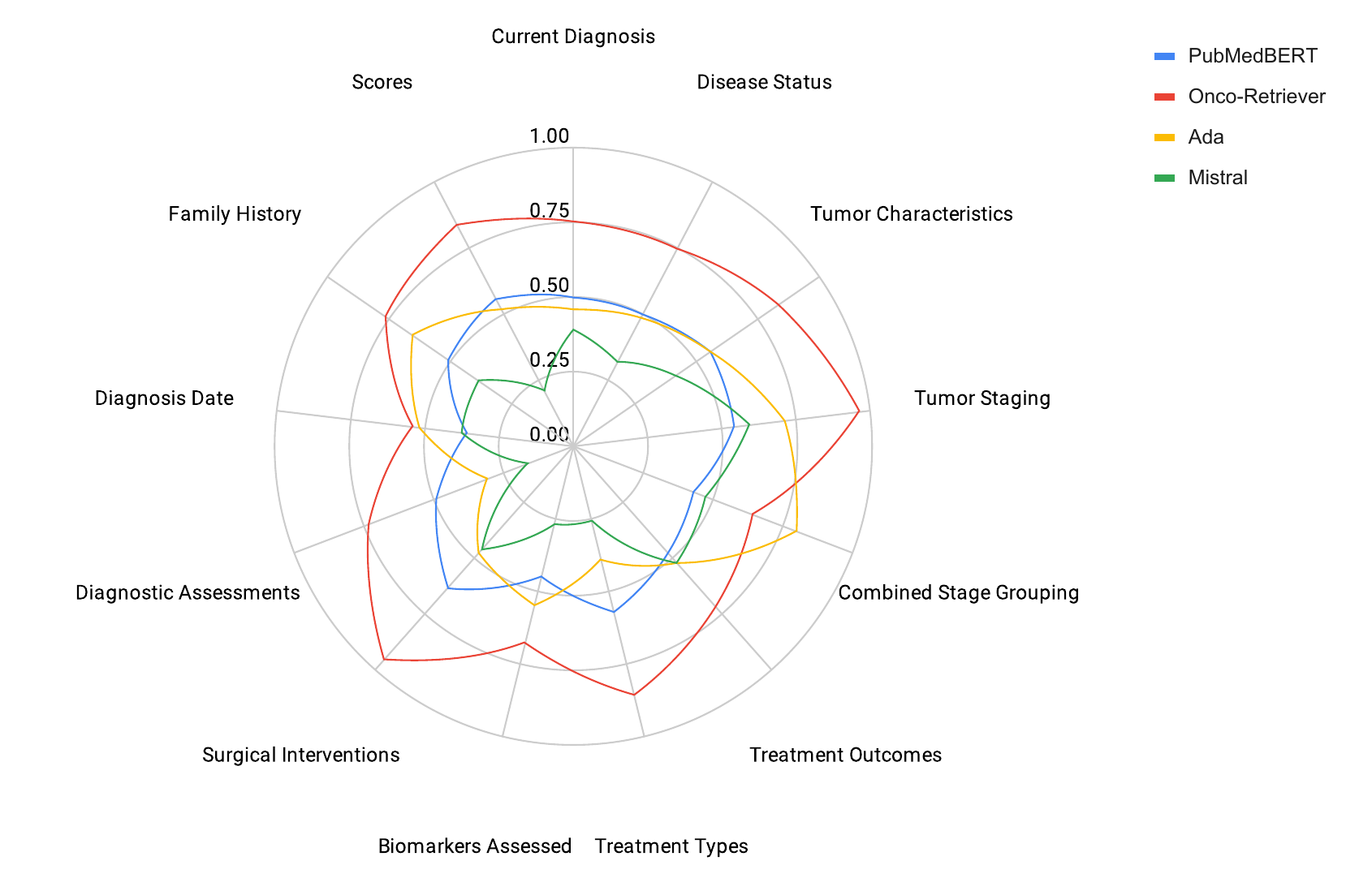}
  \caption{Normalised F1 scores across concepts for each retriever }
  \label{fig:training}
\end{figure}

Onco-Retriever (Large), built on the Gemma model with 2 billion parameters, achieved the highest overall performance, with a precision of 0.64 and recall of 0.77. The Large variant excelled in recall for most concepts, such as Tumor Staging (0.96), Surgical Interventions (0.95), and Tumor Characteristics (0.83). However, the improved performance came at the cost of increased latency (1140 seconds per patient).

\begin{figure}[ht]
  \centering
  \includegraphics[width=1\textwidth]{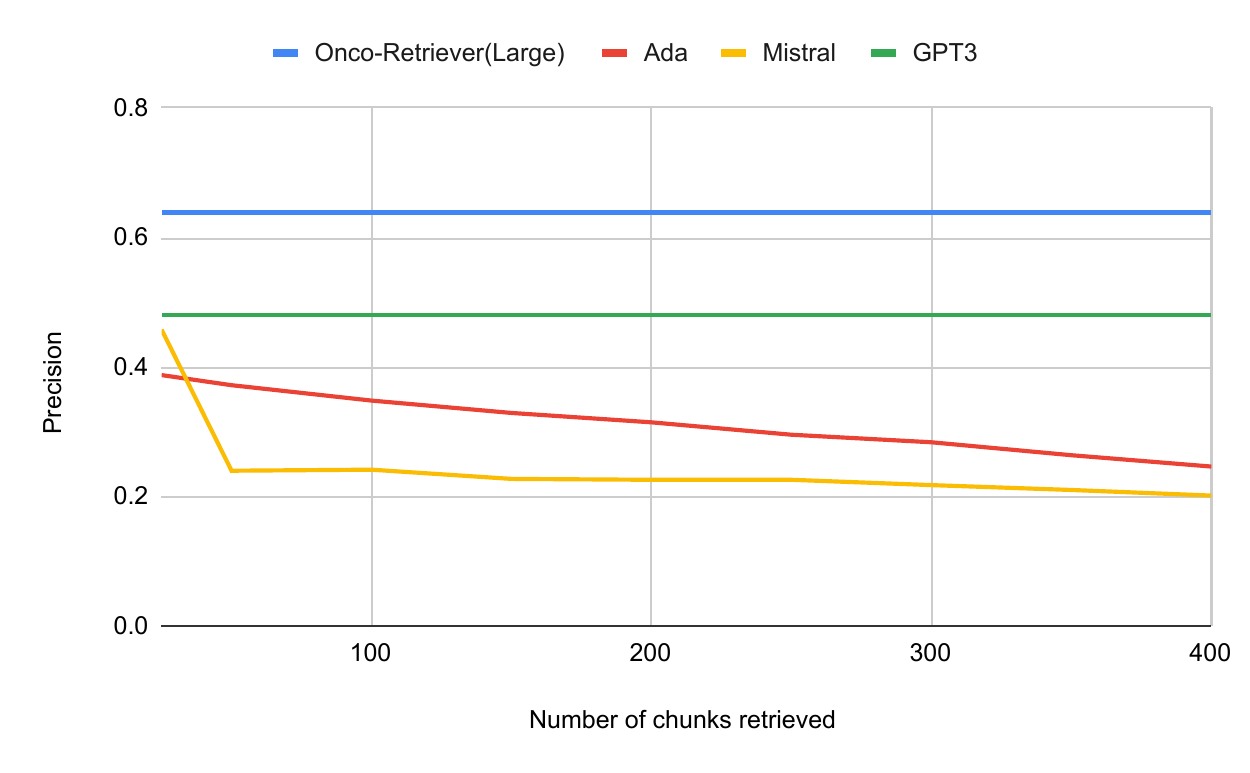}
  \caption{Precision trends with increasing nearest neighbours }
  \label{fig:training}
\end{figure}

\begin{figure}[ht]
  \centering
  \includegraphics[width=1\textwidth]{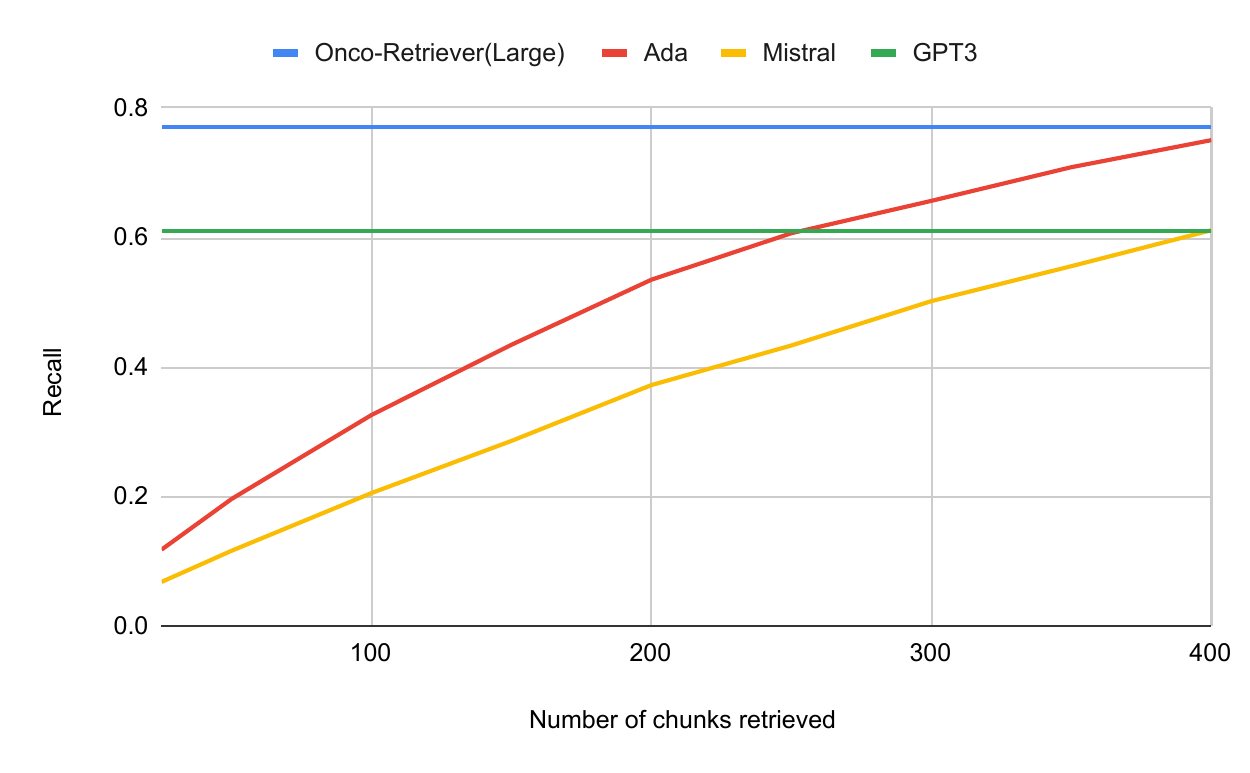}
  \caption{Recall trends with increasing nearest neighbours }
  \label{fig:training}
\end{figure}

When comparing the time taken per patient and the F1 score, Onco-Retriever (Optimized) emerged as the most efficient model, achieving an F1 score of 0.69 with a latency of only 318 seconds per patient. In contrast, Ada and Mistral had lower F1 scores (0.52 and 0.28, respectively) while requiring significantly more time per patient (2289.75 and 5160 seconds, respectively). Onco-Retriever (Small) and (Large) achieved higher F1 scores (0.67 and 0.71, respectively) but with increased latency (1200 and 1140 seconds per patient, respectively).

the performance of Ada and Mistral models varies significantly with the number of chunks retrieved. As k increases, the recall of both models improves steadily, while their precision declines. For Ada, the recall increases from 0.12 at k=25 to 0.75 at k=400, while the precision drops from 0.39 to 0.25 over the same range. Similarly, Mistral's recall improves from 0.07 at k=25 to 0.61 at k=400, with precision falling from 0.46 to 0.20.

To achieve recall levels comparable to GPT-3 and Onco-Retriever (Large), Ada and Mistral require significantly more chunks to be retrieved. For example, to reach a recall of 0.61 (matching GPT-3), Ada needs approximately 400 chunks, at which point its precision has fallen to 0.25. Mistral reaches the same recall level at k=400, with its precision dropping to 0.20.

These results demonstrate that the Onco-Retriever models, particularly the Optimized variant, provide a superior balance of performance and efficiency compared to the baseline models, making them suitable for real-world applications in oncology EHR information retrieval.




\section{Discussion} 
In our series of experiments, the Onco-Retriever demonstrated a superior performance, not only outperforming models based on embeddings but also surpassing larger classification and generative models in effectiveness. A standout feature of Onco-Retriever is its relatively smaller size, which brings considerable practical benefits. Its compact architecture allows it to operate efficiently on CPU environments, significantly reducing deployment costs while maintaining decent throughput. Furthermore, this small footprint makes the Onco-Retriever an ideal candidate for on-premise deployment. Such a feature is very important in healthcare settings, as it directly addresses concerns regarding data privacy and security.

Beyond its direct performance metrics, Onco-Retriever contributes to a broader narrative in healthcare data processing. We present robust evidence supporting the utility of large language models, like GPT-3.5, in generating synthetic yet accurately annotated datasets. This is particularly valuable in fields like healthcare, where annotated data is scarce and often difficult to obtain. The ability to create such datasets efficiently opens new avenues for advanced model training and application in data-sensitive areas.

Moreover, our approach to problem formulation distinctly deviates from traditional retriever models. We have tailored the Onco-Retriever specifically to cater to a range of practical use cases in the oncology domain, with a strong focus on enhancing summarization capabilities. This design choice not only aligns with the specific needs of oncology data processing but also illustrates the model’s adaptability to specialized healthcare contexts. 

While the current approach was confined to well-defined concepts, we plan to expand the same framework to train a general-purpose EHR query retriever, enabling successful searches across the broader EHR for a variety of use cases.

In this study, we not only explore the accuracy but also the throughput such models and found Onco-Retriever to be in the acceptable limits for processing data.


\paragraph{Limitations}
Our proposed Onco-Retriever, while demonstrating promising results, is subject to several limitations. One key limitation is its specificity to only well-defined oncology concepts; the current model is not designed for generic retrieval queries. While there is potential to adapt our approach to encompass generic queries, we have not yet validated whether the observed performance gains would be analogous in a broader context. Additionally, the task of creating a comprehensive dataset that adequately represents the vast spectrum of possible generic queries presents a difficult challenge. Our focus on retrieval of concepts that have commonalities across oncology-specific vocabularies allows a framework for expansion to capture the most high-priority queries asked in the cancer domain. This expansion is not trivial and demands more planning and execution.

Moreover, our system, despite its advancements, is not infallible. The current recall rate hovers around 78\%, which, while impressive, is not perfect. In a real-world healthcare setting, any failure to capture key information could have significant consequences. Our evaluation did not extend to integrating the retriever within a complete Retrieval-Augmented Generation (RAG) pipeline or in a real-world search use case. Such an assessment might provide deeper insights into the impact of the 20\% information potentially missed by our model. It is important to note that key medical information often recurs across patient notes. Therefore, even if our model overlooks certain details in one instance, the comprehensive nature of the input might compensate for this by capturing the required information in other snippets or chunks.

Additionally, our latency analysis indicates that while the Onco-Retriever operates with decent speed when running in the background, it is not optimized for real-time execution. This presents a significant hurdle for deployment in scenarios where immediate retrieval is important, such as during patient consultations or in real-time decision-making contexts. This limitation underscores the need for further optimization to make the Onco-Retriever a viable tool for real-time clinical use


\bibliography{references}

\end{document}